\documentclass[11pt,a4paper]{article}
\usepackage{authblk}
\usepackage[hyperref]{ranlp2023}
\usepackage{times}
\usepackage{latexsym}

\usepackage{microtype}

\usepackage{times}
\usepackage{latexsym}
\usepackage{graphicx}

\usepackage{adjustbox}
\usepackage{booktabs}
\usepackage{arydshln}
\usepackage{multirow}
\usepackage{dirtytalk}
\usepackage{color}
\usepackage{caption}
\usepackage{placeins}
\usepackage{amsmath, amssymb}
\usepackage{placeins}

\DeclareCaptionType{exam}[Example][List of Examples]

\makeatletter
\newcommand\footnoteref[1]{\protected@xdef\@thefnmark{\ref{#1}}\@footnotemark}
\makeatother

\newcommand\blfootnote[1]{%
  \begingroup
  \renewcommand\thefootnote{}\footnote{#1}%
  \addtocounter{footnote}{-1}%
  \endgroup
}

\makeatletter
\def\adl@drawiv#1#2#3{%
        \hskip.5\tabcolsep
        \xleaders#3{#2.5\@tempdimb #1{1}#2.5\@tempdimb}%
                #2\z@ plus1fil minus1fil\relax
        \hskip.5\tabcolsep}
\newcommand{\cdashlinelr}[1]{%
  \noalign{\vskip\aboverulesep
           \global\let\@dashdrawstore\adl@draw
           \global\let\adl@draw\adl@drawiv}
  \cdashline{#1}
  \noalign{\global\let\adl@draw\@dashdrawstore
           \vskip\belowrulesep}}
\makeatother

\aclfinalcopy 
\def\aclpaperid{75} 

\title{Improving Aspect-Based Sentiment with\\ End-to-End Semantic Role Labeling Model}

\author[*]{\bf  Pavel P\v{r}ib\'{a}\v{n}}
\author[*]{\bf Ond\v{r}ej Pra\v{z}\'{a}k}
\affil[ ]{University of West Bohemia, Faculty of Applied Sciences, Czech Republic}
\affil[]{Department of Computer Science and Engineering,}
\affil[]{NTIS -- New Technologies for the Information Society,}
\affil[  ]{\tt	\{pribanp,ondfa\}@kiv.zcu.cz}
\affil[  ]{\tt http://nlp.kiv.zcu.cz}

\date{}

\begin{document}
\maketitle
\begin{abstract}
This paper presents a series of approaches aimed at enhancing the performance of Aspect-Based Sentiment Analysis (ABSA) by utilizing extracted semantic information from a Semantic Role Labeling (SRL) model.
We propose a novel end-to-end Semantic Role Labeling model that effectively captures most of the structured semantic information within the Transformer hidden state. We believe that this end-to-end model is
well-suited
for our newly proposed models that incorporate semantic information. We evaluate the proposed models in two languages, English and Czech, employing ELECTRA-small models. Our combined models improve ABSA performance in both languages. Moreover, we achieved new state-of-the-art results on the Czech ABSA.

\end{abstract}

\section{Introduction}
\label{sec:intro}
In recent years, the pre-trained BERT-like models based on the Transformer \cite{attention-vaswani-2017} architecture demonstrated their performance superiority across various natural language processing (NLP) tasks. In this paper, we study the possibility of a combination of two seemingly unrelated NLP tasks: Aspect-Based Sentiment Analysis (ABSA) and Semantic Role Labeling (SRL). We believe that the structured semantic information of a sentence extracted from an SRL model can enhance the performance of an ABSA model. We investigate our assumption on the ELECTRA \cite{clark2020electra} model architecture since it is a lighter and smaller alternative to the popular and commonly used models such as BERT \cite{devlin-etal-2019-bert} or RoBERTa \cite{liu2019roberta}. Because the ELECTRA model is smaller in terms of the number of parameters, it does require less GPU memory and time to be fine-tuned.

\par Sentiment analysis (SA) is an essential part of NLP. The most prevalent SA task is the \textit{Sentiment Classification}, where the objective is to classify a text fragment (e.g., sentence or review) as \textit{positive} or \textit{negative}, eventually as \textit{neutral}. In this type of task, we assume that there is only one opinion in the text. In reality, as illustrated in Figure \ref{fig:absa-example}, this assumption often does not hold true \cite{liu2012sentiment}.

\blfootnote{
    %
    %
    %
    %
    \hspace{-0.18cm}\textsuperscript{*}Equal contribution.}

\begin{itemize}
	\small
	\item[] \hspace{-0.8cm} { \say{\textit{The \textcolor{green}{burger} was excellent but the \textcolor{red}{waitress}  was unpleasant}}}
	    \vspace{-0.27cm}
	\item[] CE $\Rightarrow$  food, service \\
	CP $\Rightarrow$  food:\textit{positive}, service:\textit{negative}
	\label{fig:absa-example}
	\captionof{figure}{Example of CE and CP subtasks of ABSA.}
\end{itemize}

Aspect-Based Sentiment Analysis \cite{liu2012sentiment,pontiki-etal-2014-semeval} focuses on detecting aspects (e.g., food or service in the restaurant reviews domain) and determining their polarity, enabling more detailed analysis and understating of the expressed sentiment. As shown by \citet{pontiki-etal-2014-semeval}, the ABSA task can be further divided into four subtasks: \textit{Aspect term extraction} (TE), \textit{Aspect term polarity} (TP), \textit{Aspect category extraction} (CE), and \textit{Aspect category polarity} (CP).

We aim at the CE and CP subtasks,\footnote{See \cite{pontiki-etal-2014-semeval} for a detailed description of all the subtasks.} and we treat them as a single classification task, see Section \ref{sec:absa-model}. 
As depicted in Figure \ref{fig:absa-example}, the goal of the CE subtask is to detect a set of aspect categories within a given sentence, 
i.e., for a given text $S = \{w_1, w_2, \dots w_n\}$ assign set $M = \{a_1, a_2, \dots, a_m \}$ of $m$ aspect categories, where $m \in [{0,k}]$, $M \subset A$ and $A$ is a set of $k$ predefined aspect categories $A = \{a_1, a_2, \dots, a_k \}$. The goal of CP is to assign one of the predefined polarity labels $p$ for each of the given (or predicted) aspect categories of the set $M$ for the given text $S$, where $p \in P = \{positive, negative, neutral\}$.

\begin{figure}[ht!]
\centering\includegraphics[width=.8\linewidth]{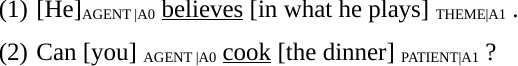}
\caption{Examples of SRL annotations.}\label{fig:srl-example}
\end{figure}

The Semantic Role Labeling task \cite{gildea2002} belongs among shallow semantic parsing techniques. The SRL goal is to identify and categorize semantic relationships or \textit{semantic roles} of given \textit{predicates}. Verbs, such as ``believe'' or ``cook'', are natural predicates, but certain nouns are also accepted as predicates. The simplified definition of semantic roles is that semantic roles are abstractions of predicate arguments. For example, the semantic roles for ``believe'' can be \textit{Agent} (a believer) and \textit{Theme} (a statement) and for ``cook'' \textit{Agent} (a chef),  \textit{Patient} (a food), \textit{Instrument} (a device for cooking) -- see examples in Figure \ref{fig:srl-example}.
The theory of predicates and their roles is very well established in several linguistic resources such as PropBank \cite{Palmer:2005} or FrameNet \cite{Baker:1998}.

\par In this work, we introduce a novel end-to-end SRL model that offers enhanced compatibility with other NLP tasks. Unlike other BERT-based models \cite{shi2019simple, papay2021constraining}, our proposed approach integrates the complete semantic information into the hidden state of the Transformer.
This end-to-end SRL model is particularly well-suited for combination with the Aspect-Based Sentiment Analysis task, as it encapsulates the entire predicate-argument structure of the sentence within a single hidden state, in contrast to the approach of \cite{shi2019simple}, which encodes each argument separately and requires gold arguments on input. Our model, on the other hand, only requires the input text.

\par We assume that leveraging the syntax and semantic information extracted from SRL can significantly enhance the performance of the aspect category polarity subtask.  This assumption is grounded in the notion that the SRL information has the potential to unveil valuable and pertinent relations between entities within a given sentence, which play a crucial role in accurate aspect category polarity predictions. This holds particularly true for longer and more complex sentences, where a broader contextual understanding becomes essential. For a concrete illustration, please refer to Appendix \ref{sec:semantic-parse-examples}.


To combine the SRL and ABSA models effectively, we propose three different approaches. Through their integration, we demonstrate performance improvements on the ABSA task for both English and Czech languages, employing ELECTRA-small models.
Moreover, we achieved new state-of-the-art (SotA) results on the Czech ABSA task.
We publicly release our source codes\footnote{\label{note:github}\url{https://github.com/pauli31/srl-aspect-based-sentiment}}.  

\section{Related Work}
\par The early studies \cite{hu-mining-2004-ABSA,Ganu2009BeyondTS,kiritchenko-etal-2014-nrc,hercig2016unsupervised} focusing on the English ABSA task relied on word n-grams, lexicons, and other feature extraction techniques in combination with supervised machine learning algorithm such as support vector machine classifiers.
These approaches were surpassed by deep neural network (DNN) models \cite{tang-etal-2016-effective,ma2017interactive,chen-etal-2017-recurrent,fan-etal-2018-multi} that typically employed recurrent neural network e.g., Long Short-Term Memory (LSTM) \cite{hochreiter1997long}.

\par Recently, the BERT-like models were successfully applied to the ABSA task.
\citet{sun-etal-2019-utilizing} solve the CE and CP subtasks at once by introducing auxiliary sentences and transforming the problem to a sentence-pair classification task. \citet{xu-etal-2019-bert} and \citet{rietzler-etal-2020-adapt} improved results by pre-training the model on the task domain data. \citet{liu-etal-2021-solving} treated the ABSA task as a text generation task outperforming the previous SotA results. \citep{zhang-etal-2019-aspect,LIANG2022107643} employed graph convolutional networks. Another related work can be found in \cite{li-etal-2020-multi-instance}.

 In \cite{sido-etal-2021-czert, priban-steinberger-2021-multilingual,lehecka-sentiment,priban-steinberger-2022-czech} the BERT-like models were used for sentiment classification and subjectivity classification, to the best of our knowledge, there is no application of BERT-like models for ABSA in the Czech language. \citet{steinberger-etal-2014-aspect} introduced the first Czech ABSA dataset from the restaurant reviews domain. They used a Maximum Entropy classifier and Conditional Random Fields for their baselines. \citet{hercig2016unsupervised} extended this dataset and improved the baseline by adding semantic features.
\citet{lenc2016neural} applied a convolutional neural network for the CP task and RNN  for the CP task to the dataset from \citet{hercig2016unsupervised}.

The pioneered approaches 
of the SRL \cite{gildea2002} task used standard feature engineering methods \cite{moschitti2008tree}. Since SRL is closely bounded with syntax, adding syntactic information is very helpful. In 2008 CoNLL shared task \cite{surdeanu2008conll} syntax-based SRL task was proposed.

In more recent years (with DNNs), the attention was drawn back to standard span-based SRL, where we form SRL as (linear) tagging. Many approaches are based on LSTMs \cite{he2017deep}.  
Later, \citet{tan2018deep}, inspired by the Transformer, proposed a self-attention-based model.

Several end-to-end models for all SRL subtasks were also introduced. \citet{he2018jointly} abandon the BIO tagging scheme, and they are rather predicting predicate-argument span tuples by searching through the possible combinations. They use a multi-layer bi-LSTM to produce contextualized representations of predicates and argument spans. 
The most recent approaches use BERT-like pre-trained models.
\citet{shi2019simple} proposed a simple BERT approach for argument identification and classification. This means, in their setting, the gold predicates are known. \citet{papay2021constraining} propose regular-constrained conditional random fields (CRF) decoding on top of the same model. There are many other complex deep models \cite{zhang2021semantic, wang2021mrc}

For our experiments, we need an end-to-end SRL model which encodes most of the information in the Transformer's hidden state. However, to the best of our knowledge, there is no such model. As a result, we introduce our end-to-end model later in this paper to fulfil this need.

Various approaches have been made to enhance one task through the integration of another, usually using multi-task learning techniques. \citet{hashimoto2016joint} proposed a joint model for learning the whole NLP stack (POS tagging, chunking, parsing, semantic relatedness, entailment). They train a single model for all tasks in a sequence (chunking after POS tagging etc.). At each layer (for each task), they use regularization on the difference from previous layer weights. They show that the tasks help each other significantly.

\par \citet{absa-aux} use dependency neighbourhood prediction and part-of-speech tagging as auxiliary tasks for ABSA. They introduced the new dependency neighbourhood prediction task to utilize the syntactic dependency information to improve the performance of the sentiment classification task. They train the auxiliary tasks together with the main sentiment classification task. The task classifies each token as either in the dependency neighbourhood or not. The dependency neighbourhood for a given token in a sentence is defined as the tokens in the sentence that are linked to the given token through, at most, $n$-hop dependency relations. \citet{zhang2020semantics} pretrain BERT model on semantic role labeling task and show, that the pretraining helps for many natural language understanding tasks. These examples of multi-task learning demonstrate the potential benefits of incorporating additional tasks in NLP models.

\section{Models}

To find an effective way to combine the models, we first fine-tune the individual models separately to find the optimal set of hyper-parameters for individual tasks. Moreover, we need SRL fine-tuned model as the input for the combined models. For ABSA, we adopt the model proposed by \cite{sun-etal-2019-utilizing}. We propose a new SRL end-to-end model, specifically designed for seamless integration with other tasks.


\subsection{Semantic Role Labeling}
Our goal is to train a universal encoder that effectively captures SRL information from a plain-text input. To accomplish this, we propose an end-to-end model with a single projection layer on the top of the ELECTRA encoder (or any other pre-trained language model). This way, all the information useful to predict role labels is encoded in the last hidden state of the encoder. Consequently, we can use this representation in other tasks. Although our end-to-end model exhibits lower performance than the commonly used BERT SRL model \cite{shi2019simple, sido-etal-2021-czert}, we believe it is more suitable for this task. 

\par In our end-to-end model, we first encode the whole sentence and then iterate over all possible word pairs (the first word is a potential predicate and the second is a potential argument). For each potential predicate-argument pair, we first concatenate the representations of predicate and argument and then classify the argument role. If the potential predicate is not a real predicate word or the potential argument is not an argument of the predicate, the role of the pair is set to \textit{Other}. If a word is represented by multiple subword tokens, only the first token is classified. This is common practice in tagging tasks where the model learns to encode the semantics of a multi-token word into the first subword, then each word has a single token on the output for its classification.


\par
\begin{figure}
    \centering
    \includegraphics[width=\linewidth]{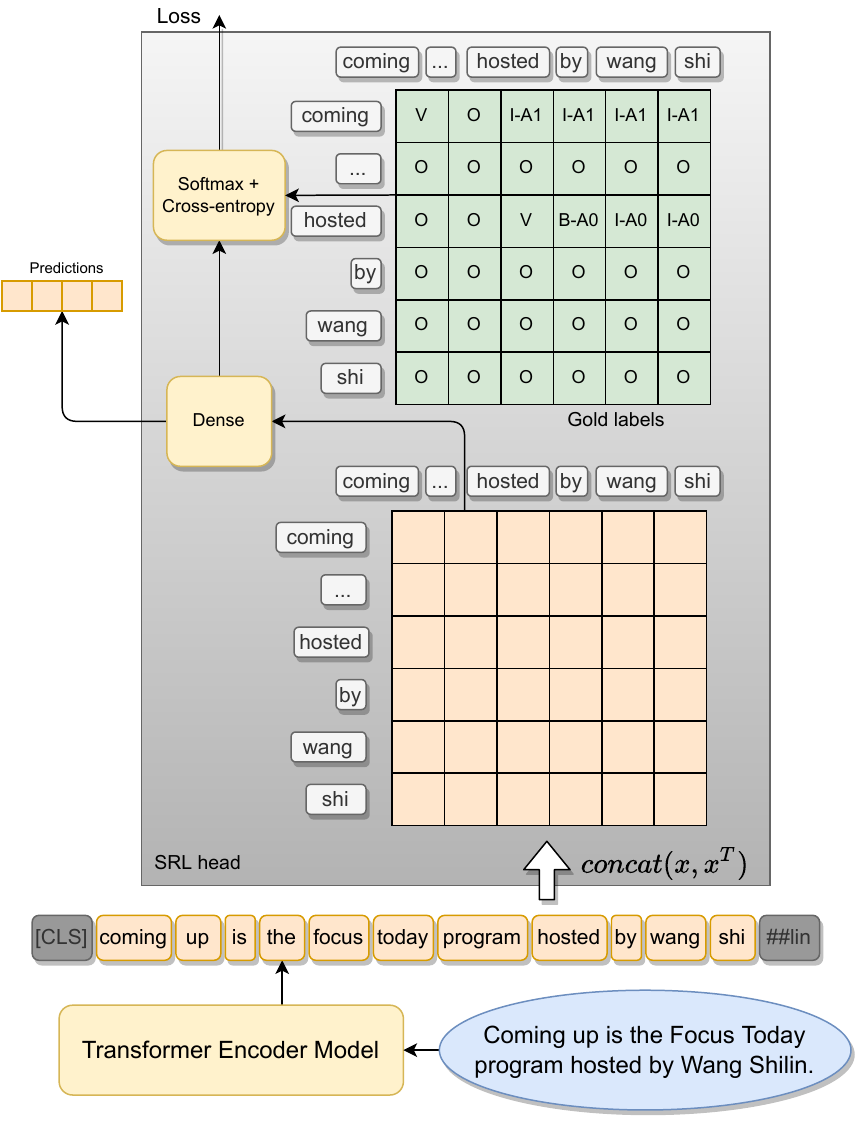}
    \caption{End-to-end SRL model architecture.}
    \label{fig:SRL}
\end{figure}

 Our approach differs from that of \citet{zhang2021semantic} in terms of how the predicate-argument structure of the sentence is encoded within the transformer model. While \citet{zhang2021semantic} encodes each argument separately and requires gold arguments on input, our model only requires plain text as input. In other words, our model requires only text as input, but the model proposed by \citet{zhang2021semantic} operates on pairs of text-predicate, producing representations solely for the input pair rather than the entire SRL output encompassing all predicates within the sentence. Figure \ref{fig:SRL} shows the schema of our end-to-end SRL model.

For our approach, it is necessary to have the same format of input (i.e., plain text) for both tasks that are combined. This is the reason why we need our end-to-end SRL model.
For multitask learning, we need a general-purpose model, the same for both tasks. The task-specific models may yield better results on the SRL task, but they are specifically oriented only on the SRL task and makes their integration with ABSA or utilization in multitask learning challenging, if not impossible.

\subsection{Aspect-Based Sentiment}
\label{sec:absa-model}
\par As we mentioned in the introduction, we tackle the CE and CP subtasks of ABSA, as one classification task. We adopt the same approach as \citet{sun-etal-2019-utilizing}, and we construct auxiliary sentences and convert the subtasks to a binary classification task.

We use the NLI-B approach from \citet{sun-etal-2019-utilizing} to build the auxiliary sentences. For each sentence, we build multiple auxiliary pseudo sentences that are generated for every combination of all polarity labels and aspect categories\footnote{For English we have four polarity labels plus artificial label \textit{none} and five aspect categories, i.e. $25$ possible auxiliary sentences. For Czech there is $20$ possible sentences ($3+1$ polarity labels and five aspect categories).}. Each example has a binary label $l \in \{0, 1\}$; $l = 1$ if the auxiliary sentence corresponds to the original labels, $l = 0$ otherwise. We also add the artificial polarity class \textit{none} that has assigned binary label $l = 1$ if there is no aspect category for a given sentence. The pseudo auxiliary sentence consists only of a polarity label and aspect category in a given language. For example, the auxiliary sentences for all aspects of the sentence \say{\textit{The \textcolor{green}{burger} was excellent but the \textcolor{red}{waitress}  was unpleasant}} are shown in Figure \ref{fig:absa-example-appendix}.




\vspace{0.4cm}
\begin{adjustbox}{width=0.9\linewidth,center}

\begin{tabular}{cllccll} \toprule
\multicolumn{1}{c}{label}  &  & \multicolumn{1}{c}{sentence}&  & \multicolumn{1}{l}{label} &  & \multicolumn{1}{c}{sentence} \\ \midrule
\multicolumn{3}{c}{food} & & \multicolumn{3}{c}{service} \\ \cline{1-3} \cline{5-7}
1 & $\Rightarrow$ & positive -- food & & 0 & $\Rightarrow$ & positive -- service \\
0 & $\Rightarrow$ & negative -- food & & 1 & $\Rightarrow$ & negative -- service \\
0 & $\Rightarrow$ & neutral -- food & & 0 & $\Rightarrow$ & neutral -- service \\
0 & $\Rightarrow$ & conflict -- food & & 0 & $\Rightarrow$ & conflict -- service \\
0 & $\Rightarrow$ & none -- food & & 0 & $\Rightarrow$ & none -- service \\ \cdashlinelr{1-7}
\multicolumn{3}{c}{price} & & \multicolumn{3}{c}{ambience} \\ \cline{1-3} \cline{5-7}
0 & $\Rightarrow$ & positive -- price & & 0 & $\Rightarrow$ & positive -- ambience \\
0 & $\Rightarrow$ & negative -- price & & 0 & $\Rightarrow$ & negative -- ambience \\
0 & $\Rightarrow$ & neutral -- price & & 0 & $\Rightarrow$ & neutral -- ambience \\
0 & $\Rightarrow$ & conflict -- price& & 0 & $\Rightarrow$ & conflict -- ambience \\
1 & $\Rightarrow$ & none -- price & & 1 & $\Rightarrow$ & none -- ambience \\ \cdashlinelr{1-7}
\multicolumn{3}{c}{general} & & \multicolumn{1}{l}{} &  &  \\ \cline{1-3}
0 & $\Rightarrow$ & positive -- general & & \multicolumn{1}{l}{} &  &  \\
0 & $\Rightarrow$ & negative -- general & & \multicolumn{1}{l}{} &  &  \\
0 & $\Rightarrow$ & neutral -- general  & & \multicolumn{1}{l}{} &  &  \\
0 & $\Rightarrow$ & conflict -- general & & \multicolumn{1}{l}{} &  &  \\
1 & $\Rightarrow$ & none -- general & & \multicolumn{1}{l}{} &  & \\ \bottomrule
\end{tabular}

\end{adjustbox}
	\captionof{figure}{Example of auxiliary sentences.}
	\label{fig:absa-example-appendix}
\vspace{0.4cm}

\par Each auxiliary sentence is combined with the original sentence and separated with \texttt{[SEP]} token and forms one training example, e.g., \texttt{[CLS]} \textit{positive - food} \texttt{[SEP]} \textit{the burger was excellent but the waitress was unpleasant} \texttt{[SEP]}. We fine-tune the pre-trained transformer model for the binary classification task on all generated training examples as \citet{sun-etal-2019-utilizing}.


\subsection{Combined Models}

\par
\begin{figure}[ht!]
    \centering
    \includegraphics[width=.6\linewidth]{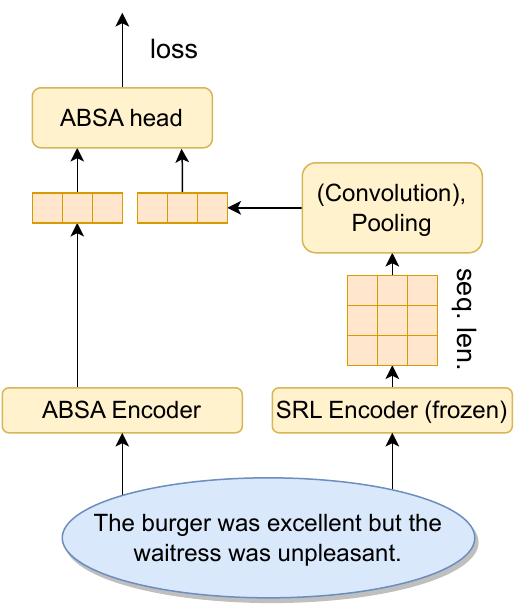}
    \caption{Concat model architecture.}
    \label{fig:concat}
\end{figure}

 We propose several models designed to use SRL representation to enhance ABSA performance. The first type of model predicts aspect and sentiment using concatenated representations from both the SRL and ABSA encoders. The SRL encoder is pre-trained (pre-fine-tuned) on the SRL data, and its weights remain fixed during sentiment training. Since SRL is a token-level task, we need to reduce the sequential dimension before performing the concatenation step. To address this, we employ two approaches: simple average-over-time pooling (named \textit{concat-avg}) and a convolution layer followed by max-over-time pooling (named \textit{concat-conv}). Figure \ref{fig:concat} shows the model architecture.

\par
\begin{figure}[ht!]
    \centering
    \includegraphics[width=.9\linewidth]{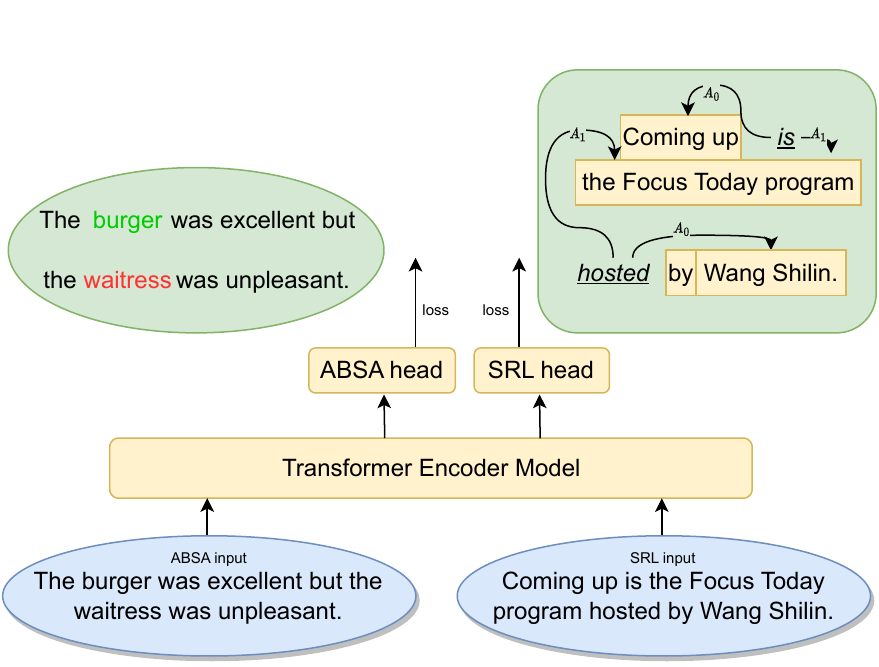}
    \caption{Multi-task model architecture.}
    \label{fig:multitask}
\end{figure}

 The last model uses standard multi-task learning. We utilize a single Transformer encoder with two classification heads: one for the sentiment (standard head for sequence classification) and the other for SRL (the head architecture is presented in the previous section with the end-to-end SRL model). The model is trained using alternating batches, it means that we use different training data for both tasks, and we are not mixing them in a batch. In a single batch, we provide only ABSA or SRL data. See Figure \ref{fig:multitask} model's architecture.

\section{Experiments}
\label{sec:experiments}
In our experiments, we aim to verify our idea that injected SRL information can improve the results of the ABSA task, particularly the CP subtask.

\begin{table*}[ht!]
    \begin{adjustbox}{width=0.8\linewidth,center}
\begin{tabular}{lllllll} \toprule
\multirow{2}{*}{Model} & \multicolumn{3}{c}{Category Extraction} &  & \multicolumn{2}{c}{Category Polarity} \\ \cline{2-4}  \cline{6-7}
 & \multicolumn{1}{c}{F1 Micro} & Precision & Recall &  & \multicolumn{1}{c}{Acc \#3} & \multicolumn{1}{c}{Acc \#2} \\ \midrule
baseline  & 86.04$^{\pm0.36}$ & 86.48$^{\pm0.97}$ & 85.62$^{\pm0.65}$ &  & 75.58$^{\pm0.55}$ & 88.69$^{\pm0.26}$ \\
concat-conv  & \underline{\textbf{86.58}}$^{\pm0.54}$ & \underline{\textbf{86.90}}$^{\pm0.51}$ & \underline{\textbf{86.28}}$^{\pm0.94}$ &  & \underline{\textbf{79.20}}$^{\pm0.48}$ & \underline{\textbf{90.26}}$^{\pm0.58}$ \\
concat-avg  & 86.34$^{\pm0.57}$ & 86.57$^{\pm0.84}$ & 86.12$^{\pm1.08}$ &  & 78.33$^{\pm0.64}$ & 90.06$^{\pm0.79}$ \\
multi-task  & 85.62$^{\pm0.63}$ & 86.24$^{\pm0.66}$ & 85.01$^{\pm0.66}$ &  & 77.27$^{\pm0.69}$ & 89.00$^{\pm0.63}$ \\ \cdashlinelr{1-7} 
baseline \cite{hercig2016unsupervised}* & 71.70\phantom{****} & \multicolumn{1}{c}{-} & \multicolumn{1}{c}{-} &  & 69.70\phantom{****} & \multicolumn{1}{c}{-} \\
best \cite{hercig2016unsupervised}* & 80.00\phantom{****} & \multicolumn{1}{c}{-} & \multicolumn{1}{c}{-} &  & 75.20\phantom{****} & \multicolumn{1}{c}{-} \\
CNN2 \cite{lenc2016neural} & \multicolumn{1}{c}{-} & \multicolumn{1}{c}{-} & \multicolumn{1}{c}{-} &  & 69.00$^{\pm2.00}$ & \multicolumn{1}{c}{-} \\ \bottomrule
\end{tabular}
\end{adjustbox}
\caption{Czech results for the category extraction (CE) subtask as F1 Micro score, Precision and Recall. Results for the category polarity (CP) subtask as accuracy for three polarity labels (Acc \#3) and binary polarity labels (Acc \#2). Results marked with * symbol were obtained by 10-fold cross-validation.} \label{tab:results-cs}
\end{table*}

\begin{table*}[ht!]
    \begin{adjustbox}{width=0.95\linewidth,center}
\begin{tabular}{llllllll} \toprule
\multicolumn{1}{c}{\multirow{2}{*}{Model}} & \multicolumn{3}{c}{Category Extraction} &  & \multicolumn{3}{c}{Category Polarity} \\ \cline{2-4}  \cline{6-8}
\multicolumn{1}{c}{} & \multicolumn{1}{c}{F1 Micro} & Precision & Recall &  & Acc \#4 & \multicolumn{1}{c}{Acc \#3} & \multicolumn{1}{c}{Acc \#2} \\ \midrule
baseline & 89.50$^{\pm0.45}$ & 90.95$^{\pm0.70}$ & 88.09$^{\pm0.48}$ &  & 83.03$^{\pm0.43}$ & 86.91$^{\pm0.55}$ & 92.74$^{\pm0.53}$ \\
concat-conv & \textbf{89.74}$^{\pm0.55}$ & \textbf{91.24}$^{\pm0.54}$ & \textbf{88.28}$^{\pm0.77}$ &  & \textbf{84.19}$^{\pm0.49}$ & \textbf{88.08}$^{\pm0.41}$ & \textbf{93.76}$^{\pm0.46}$ \\
concat-avg & 89.58$^{\pm0.43}$ & 91.15$^{\pm0.60}$ & 88.08$^{\pm0.66}$ &  & 84.13$^{\pm0.51}$ & 87.95$^{\pm0.46}$ & 93.49$^{\pm0.44}$ \\
multi-task & 89.36$^{\pm0.15}$ & 90.72$^{\pm0.52}$ & 88.05$^{\pm0.44}$ &  & 82.83$^{\pm1.10}$ & 87.05$^{\pm1.21}$ & 92.74$^{\pm0.79}$ \\ \cdashlinelr{1-8} 
XRCE \cite{brun-etal-2014-xrce}  & 82.29\phantom{****} & 83.23\phantom{****} & 81.37\phantom{****} &  & 78.10\phantom{****} & \multicolumn{1}{c}{-}\phantom{****} & \multicolumn{1}{c}{-}\phantom{****} \\
NRC \cite{kiritchenko-etal-2014-nrc}  & 88.58\phantom{****} & 91.04\phantom{****} & 86.24\phantom{****} &  & 82.90\phantom{****} & \multicolumn{1}{c}{-}\phantom{****} & \multicolumn{1}{c}{-}\phantom{****} \\
BERT single \cite{sun-etal-2019-utilizing}  & 90.89\phantom{****} & 92.78\phantom{****} & 89.07\phantom{****} &  & 83.70\phantom{****} & 86.90\phantom{****} & 93.30\phantom{****} \\
NLI-B \cite{sun-etal-2019-utilizing}  & 92.18\phantom{****} & 93.57\phantom{****} & 90.83\phantom{****} &  & 84.60\phantom{****} & 88.70\phantom{****} & 95.10\phantom{****} \\
QACG-B \cite{wu2021context}  & 92.64\phantom{****} & 94.38$^{\pm0.31}$ & \underline{90.97}$^{\pm0.28}$ &  & \underline{86.80}$^{\pm0.80}$ & 90.10$^{\pm0.30}$ & \underline{95.60}$^{\pm0.40}$ \\
BART generation \cite{liu-etal-2021-solving}  & \underline{92.80}\phantom{****} & \underline{95.18}\phantom{****} & 90.54\phantom{****} &  & \multicolumn{1}{c}{-}\phantom{****} & \underline{90.55}$^{\pm0.32}$ & \multicolumn{1}{c}{-}\phantom{****}
\\ \bottomrule
\end{tabular}
\end{adjustbox}
\caption{English results for the category extraction (CE) subtask as F1 Micro score, Precision and Recall. Results for category polarity (CP) subtask as accuracy for four polarity labels (Acc \#4), three polarity labels (Acc \#3) and binary polarity labels (Acc \#2).} \label{tab:results-en}
\end{table*}

\subsection{Datasets \& Models Fine-Tuning}

\par For Semantic Role Labeling, we use OntoNotes 5.0 dataset \cite{weischedel2013ontonotes} for English and CoNLL 2009 \cite{hajic2009conll} for Czech. As metrics, we report the whole role F1 score for both datasets. Additionally, for English, we report CoNLL 2003 official score as a comparative metric as it is the standard metric used with OntoNotes. 

\par For Aspect-Based Sentiment, we use the widely-used English dataset from \citet{pontiki-etal-2014-semeval} that consists of 3,044 train and 800 test sentences from the restaurant domain. The English dataset contains four sentiment labels: \textit{positive}, \textit{negative}, \textit{neutral}, and \textit{conflict}. Further, we split\footnote{\label{foot:github}For both English and Czech we provide a script to obtain the same split distribution.} the original training part of 3,044 sentences into development (10\%) and training parts (90\%).

For Czech experiments, we employ the dataset from \citet{hercig2016unsupervised} with 2,149 sentences from the restaurant domain. Unlike in the English dataset, there are only three polarity labels: \textit{positive}, \textit{negative}, and \textit{neutral}. Because the dataset has no official split, we divided\footref{foot:github} the data into training, development, and testing parts with the following ratio: $72 \%$ for training, $8 \%$ for the development evaluation, and $20 \%$ for testing. Both Czech and English datasets contain five aspect categories: \textit{food}, \textit{service}, \textit{price}, \textit{ambience}, and \textit{general}.

For our experiments on English, we use the pre-trained \textit{ELECTRA-small} model introduced by \citet{clark2020electra}, which has 14M parameters. For Czech, we employ the pre-trained monolingual model \textit{Small-E-Czech} \cite{kocian2021siamese} with the same size and architecture. Firstly, we train separate models for both tasks (ABSA and SRL) and select the optimal set of hyper-parameters on the development data. We then use the same hyper-parameters in combined models. For the details of hyper-parameters, see Appendix \ref{appendix:hp-params}.



\subsection{Results \& Discussion}
We report the results of our end-to-end SRL model in Table \ref{tab:e2eSRL-performance}. As we expected, our model performs worse than the model proposed by \citet{shi2019simple}, but the results are reasonably high (considering that it does not have gold predicates on input).

\begin{table}[ht!]
    \centering
    \begin{adjustbox}{width=0.9\linewidth,center}
    \begin{tabular}{cccc} \toprule
        Model & EN & EN-conll05 & CS \\ \midrule
        \cite{shi2019simple} & 88.89 & 85.20 & 83.09 \\
        end-to-end (ours) & 84.54 & 81.51 & 79.74\\ \bottomrule
    \end{tabular}
    \end{adjustbox}
    \caption{Comparison of results of the standard model and our end-to-end SRL model (reported in F1 scores, the official metrics, for the datasets used).}
    \label{tab:e2eSRL-performance}
\end{table}

Results for our ABSA experiments in Czech and English are shown in Tables \ref{tab:results-cs} and \ref{tab:results-en}, respectively. The \textit{baseline} refers to the model described in Section \ref{sec:absa-model} without any injected SRL information.

The SotA results are underlined and the best results for our experiments are bold. We include the results with the 95\% confidence interval (experiments repeated 12 times). We use the F1 Micro and accuracy for the CE and CP subtasks, respectively.

\par Based on the results presented in Tables \ref{tab:results-cs} and \ref{tab:results-en}, we can observe that our proposed models (\textit{concat-conv} and \textit{concat-avg}) with injected SRL information consistently enhance results for the CP subtask in both languages. These improvements are statistically significant. The performance of the \textit{concat-conv} and \textit{concat-avg} models does not exhibit a significant difference. In the CE subtask, we achieve the same results as the \textit{baseline} model. We think that the CE subtask is more distant from the SRL task than the CP subtask and therefore, the injection of the semantic information does not help. In other words, the semantic structure of the sentence may not play a crucial role in aspect detection (that can be viewed as multi-label text classification). On the other hand, for the CP subtask, the combined models can leverage the semantic structure of the sentence to their advantage.

\par For the Czech ABSA dataset we achieve new SotA results on both subtasks\footnote{It is worth noting that although the test data we used differ from those used by \citet{hercig2016unsupervised} due to their 10-fold cross-validation, the performance difference is substantial enough to demonstrate the superiority of our approach.}. As we expected, we did not outperform the current SotA results for the English dataset, as our ELECTRA model has considerably fewer parameters than SotA models. For Czech, the \textit{multi-task} model exhibited a marginal improvement in the results and generally, the model was significantly inferior to our other models. We decided to use the smaller ELECTRA-based models because of their much smaller computation requirements. However, in future work, we plan comparison with larger models like BERT or RoBERTa to obtain the overall performance overview of our approach.


\section{Conclusion}
In this work, we introduce a novel end-to-end SRL model that we use to improve the aspect category polarity task.  Our contribution lies in proposing several methods to integrate SRL and ABSA models, which ultimately lead to improved performance. The experimental results validate our initial assumption that leveraging semantic information extracted from an SRL model can significantly enhance the aspect category polarity task. Importantly, the approaches we propose are versatile and can be applied to combine Transformer-based models for other related tasks as well, extending the scope of their applicability.

\par Moreover, we believe that our approaches hold even greater potential in addressing other ABSA subtasks, namely term extraction and term polarity classification. These subtasks could benefit from the integration of SRL and ABSA models in a similar manner. Further, we would like to validate our approach on larger models, for example, BERT or RoBERTa.



\section*{Acknowledgements}
This work has been partly supported by grant No. SGS-2022-016 Advanced methods of data processing and analysis.
Computational resources were provided by the e-INFRA CZ project (ID:90140), supported by the Ministry of Education, Youth and Sports of the Czech Republic.


\bibliography{anthology,custom}
\bibliographystyle{acl_natbib}


\appendix

\makeatletter
\setlength{\@fptop}{0pt}
\makeatother
\begin{figure*}[t!]
    \centering
    \includegraphics[width=\linewidth]{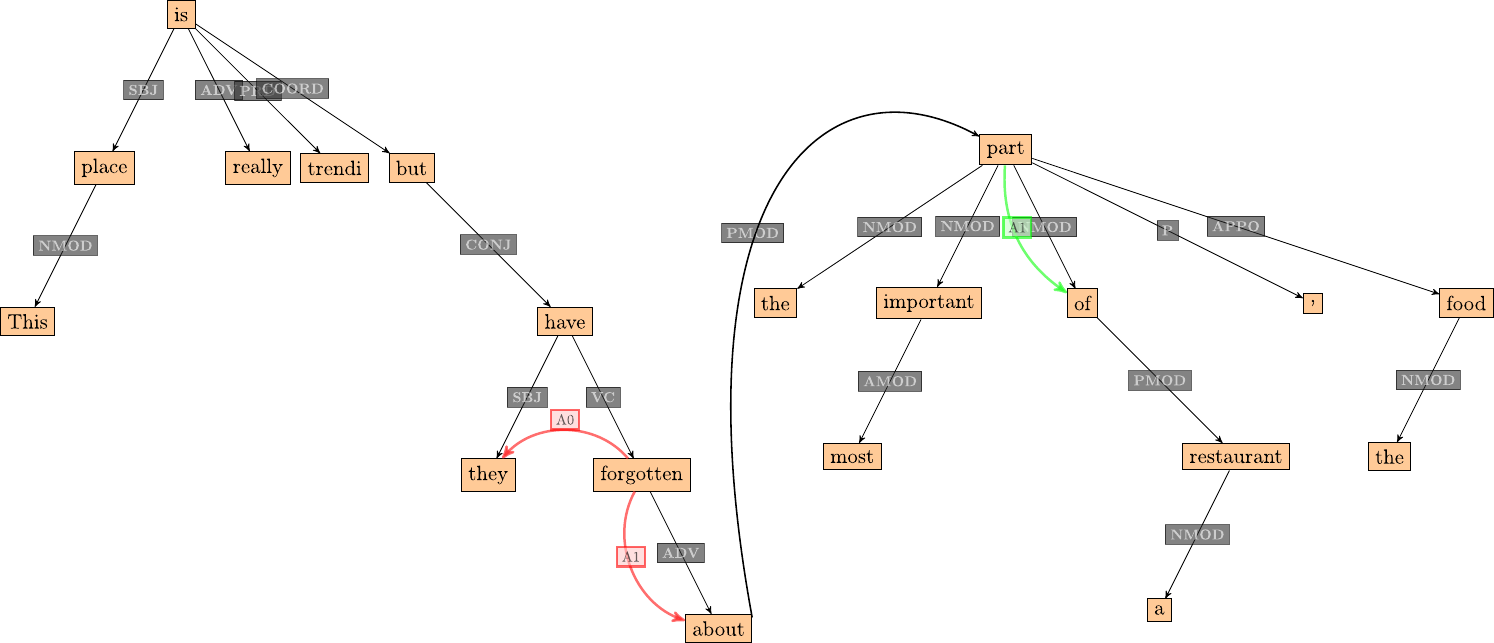}
    \caption{Example of syntactic and semantic parse tree of the following sentence \say{\textit{This place is really trendi but they have forgotten about the most important part of a  restaurant, the food}}.}
    \label{fig:srl-sentiment-tree}
\end{figure*}

\section{Training Hyper-Parameters} \label{appendix:hp-params}
We use the Adam \cite{Kingma-adam} optimizer with default parameters ($\beta_1 = 0.9, \beta_2 = 0.999$) and the cross-entropy loss function for all our experiments. The initial learning rate is set to 2e-5 with linear decay to zero. We fine-tune all models with batch size 32 and maximum sequence length 256. All data fit into this length. The models are trained for 120 epochs in Czech and 40 in English. The epochs are measured in ABSA data. The multi-task model is trained on the same amount of SRL data additionally (because we use alternating batches).

\section{Semantic Parse Tree Example}
\label{sec:semantic-parse-examples}
As mentioned in the introduction section, we assume that leveraging SRL information can prove advantageous in the aspect category polarity (CP) task. To illustrate this point, consider the annotation depicted in Figure \ref{fig:absa-example-2}, where we can observe the SRL relation extracted (see Figure \ref{fig:srl-sentiment-tree}) between the words \textit{forgotten} and \textit{food}. The information about this relation can help to understand the model that these words are related and help the model to predict the negative polarity of the food aspect category.

\newpage
\begin{itemize}
	\small
	\item[] \hspace{-0.8cm} { \say{\textit{This \textcolor{green}{place} is really trendy but they have forgotten about the most important part of a  restaurant, the \textcolor{red}{food}. }}}
	    \vspace{-0.27cm}
	\item[] CE $\Rightarrow$ food, ambience \\
	CP $\Rightarrow$  food:\textit{negative}, ambience:\textit{positive}
	\captionof{figure}{Example of CE and CP annotations.}\label{fig:absa-example-2}
\end{itemize}

\label{sec:semantic-parse-examplse}




\end{document}